\documentclass[11pt]{article}
\usepackage[utf8]{inputenc}
\usepackage[margin=1in]{geometry}
\usepackage{amsmath,amssymb,amsthm}
\usepackage{booktabs}
\usepackage{hyperref}
\usepackage{enumitem}
\usepackage{xurl}

\title{\textbf{Implementing Tensor Logic: Unifying Datalog and Neural Reasoning via Tensor Contraction}}

\author{
    Swapn Shah \\
    School of Data Science\\
    University of North Carolina at Charlotte\\
    \texttt{sshah100@charlotte.edu}
    \and
    Wlodek Zadrozny \\
    Department of Computer Science\\
    University of North Carolina at Charlotte\\
    \texttt{wzadrozn@charlotte.edu}
}

\date{}

\begin{document}

\maketitle

\begin{abstract}
The unification of symbolic reasoning and neural networks remains a central challenge in artificial intelligence. Symbolic systems offer reliability and interpretability but lack scalability, while neural networks provide learning capabilities but sacrifice transparency. Tensor Logic, proposed by Domingos \cite{domingos2025}, suggests that logical rules and Einstein summation are mathematically equivalent, offering a principled path toward unification. This paper provides empirical validation of this framework through three experiments. First, we demonstrate the equivalence between recursive Datalog rules and iterative tensor contractions by computing the transitive closure of a biblical genealogy graph containing 1,972 individuals and 1,727 parent-child relationships, converging in 74 iterations to discover 33,945 ancestor relationships. Second, we implement reasoning in embedding space by training a neural network with learnable transformation matrices, demonstrating successful zero-shot compositional inference on held-out queries. Third, we validate the Tensor Logic superposition construction on FB15k-237, a large-scale knowledge graph with 14,541 entities and 237 relations. Using Domingos's relation matrix formulation $R_r = E^\top A_r E$, we achieve MRR of 0.3068 on standard link prediction and MRR of 0.3346 on a compositional reasoning benchmark where direct edges are removed during training, demonstrating that matrix composition enables multi-hop inference without direct training examples.
\end{abstract}

\section{Introduction}
Can symbolic reasoning and neural learning be unified under a single mathematical framework? Domingos \cite{domingos2025} proposes Tensor Logic as an answer, observing that Datalog rules and Einstein summation are fundamentally the same operation. This paper provides the first empirical validation of that framework, demonstrating through three experiments that tensor operations can exactly implement logical inference and that learned embeddings compose correctly for multi-hop reasoning. We also provide the first open-source implementation of Tensor Logic with reproducible experimental traces.

Artificial intelligence research has historically followed two divergent paths. The symbolic tradition, exemplified by logic programming languages like Prolog and Datalog, emphasizes precise reasoning over structured knowledge representations. These systems excel at tasks requiring logical inference, explanation generation, and adherence to formal constraints. However, they struggle with noisy real-world data, require manual knowledge engineering, and face scalability limitations when applied to large domains. It is important to note that Datalog represents a restricted fragment of logic; it is polynomial-time decidable, whereas first-order logic is undecidable. Our validation focuses specifically on Datalog-style reasoning, not the full breadth of symbolic logic.

The connectionist tradition, now dominated by deep learning, takes the opposite approach. Neural networks learn distributed representations from raw data and have achieved remarkable success in perception tasks including image recognition, speech processing, and natural language understanding. Yet these systems operate as black boxes, offering limited insight into their decision-making processes. They also struggle with systematic generalization, often failing on compositional tasks that require combining learned concepts in novel ways.

The tension between these paradigms has motivated decades of research into neuro-symbolic integration. Early efforts attempted to extract symbolic rules from trained neural networks or to use neural networks as components within larger symbolic systems. More recent approaches have sought tighter integration, embedding logical constraints into neural architectures or learning logical rules through differentiable relaxations \cite{manhaeve2018,rocktaschel2017}. Despite significant progress, most neuro-symbolic systems still treat the neural and symbolic components as separate modules that must be carefully interfaced, limiting the benefits of true unification.

Domingos \cite{domingos2025} proposes Tensor Logic as a solution to this integration problem. The key insight is that Datalog rules and Einstein summation are fundamentally the same operation, differing only in their atomic data types. A logical rule performs joins and projections over Boolean tensors, while a neural network performs the same operations over real-valued tensors. By recognizing this equivalence, Tensor Logic provides a unified mathematical foundation where symbolic reasoning and neural learning can coexist within a single computational framework.

This paper provides empirical validation of the Tensor Logic framework through three complementary experiments:

\begin{enumerate}[leftmargin=*, itemsep=0.5em]
    \item \textbf{Symbolic Transitive Closure:} We validate that recursive Datalog rules for computing transitive closure can be exactly implemented as iterative matrix multiplication with a Heaviside step function. Using biblical genealogy data, we demonstrate that the tensor-based approach correctly discovers all ancestor relationships while preserving logical consistency.
    
    \item \textbf{Embedding Space Reasoning:} We validate that relations can be learned as transformation matrices in embedding space, enabling compositional inference through chained matrix multiplication. Using geographic data, we show that a model trained only on single-hop facts can correctly answer multi-hop queries through tensor contraction.
    
    \item \textbf{Knowledge Graph Link Prediction and Compositional Reasoning:} We validate the Tensor Logic superposition construction on FB15k-237, implementing Domingos's formulation where each relation matrix is derived from graph facts as $R_r = E^\top A_r E$. We evaluate both standard link prediction and a compositional reasoning benchmark where direct edges are removed to test whether matrix composition $e_a \cdot R_{r_1} \cdot R_{r_2}$ enables multi-hop inference.
\end{enumerate}

Together, these experiments demonstrate that Tensor Logic successfully bridges the symbolic-neural divide, providing a practical framework for unified reasoning and learning.

\section{Literature Review}

\subsection{Foundations of Logic Programming}

Logic programming emerged in the 1970s as an approach to computation based on formal logic. The theoretical foundations established that Horn clauses could be given both a declarative reading as logical statements and a procedural reading as computation rules. This dual interpretation became the basis for Prolog, one of the first practical logic programming languages.

Datalog, a restricted subset of Prolog without function symbols, gained prominence in database theory. Its limited expressiveness guarantees termination and enables efficient bottom-up evaluation strategies. A Datalog program consists of facts and rules. Facts assert ground atoms such as $\text{Parent}(\text{Alice}, \text{Bob})$, stating that Alice is a parent of Bob. Rules define derived relations in terms of existing ones. For example, the transitive closure rule for ancestry is:
\begin{equation}
\text{Ancestor}(x, z) \leftarrow \text{Ancestor}(x, y) \wedge \text{Parent}(y, z)
\end{equation}

This rule states that $x$ is an ancestor of $z$ if $x$ is an ancestor of some $y$ who is a parent of $z$. Combined with a base case stating that parents are ancestors, repeated application of this rule computes the full transitive closure of the parent relation.

Lloyd \cite{lloyd1987} provided the definitive mathematical treatment of logic programming semantics, establishing the theoretical foundations for fixpoint computation, negation as failure, and the relationship between declarative and procedural semantics. These foundations remain essential for understanding how logical inference can be mapped to computational operations.

\subsection{Statistical Relational Learning}

The desire to handle uncertainty in logical systems led to statistical relational learning, which combines logic programming with probabilistic graphical models. Markov Logic Networks, introduced by Richardson and Domingos \cite{richardson2006}, attach weights to first-order logic formulas. A formula with high weight represents a strong constraint, while a formula with low weight represents a weak preference. The weights define a probability distribution over possible worlds through a log-linear model.

Markov Logic Networks demonstrated that probabilistic and logical reasoning could be unified within a single framework. However, inference in these models requires computing partition functions over exponentially large state spaces, limiting scalability to domains with relatively few ground atoms.

\subsection{Neuro-Symbolic Integration}

The limitations of purely symbolic and purely neural approaches have motivated extensive research into hybrid systems. Logic Tensor Networks, introduced by Serafini and Garcez \cite{serafini2016}, embed first-order logic formulas into neural network architectures. Logical constants are represented as feature vectors, and logical connectives are implemented through neural network layers. The system learns embeddings that satisfy logical constraints expressed as differentiable loss functions.

Logical Neural Networks from IBM Research \cite{riegel2020} take a different approach, creating a one-to-one correspondence between neural network structure and logical formulas. Each neuron represents a logical operation, and the network architecture mirrors the structure of the knowledge base. This design enables direct interpretation of network behavior in logical terms while maintaining differentiability for learning.

Neural Theorem Provers \cite{rocktaschel2017} replace symbolic unification with differentiable operations using radial basis function kernels. This allows end-to-end training of theorem proving systems through backpropagation. The approach demonstrates that core logical operations can be made differentiable without sacrificing their semantic content.

DeepProbLog \cite{manhaeve2018} extends probabilistic logic programming with neural predicates. Neural networks provide probability distributions over ground atoms, which are then used within a probabilistic logic program. The system supports end-to-end training where gradients flow through both the neural and symbolic components.

Despite these advances, most neuro-symbolic systems maintain a separation between neural and symbolic components. For a comprehensive survey of neuro-symbolic integration approaches, see Lappin \cite{lappin2025}. Tensor Logic offers a more fundamental unification by showing that symbolic and neural operations are mathematically equivalent.

\subsection{Knowledge Graph Embeddings}

Knowledge graph embedding methods learn distributed representations of entities and relations that support reasoning through vector operations. These methods provide important precedents for Tensor Logic's approach to reasoning in embedding space.

TransE \cite{bordes2013} introduced the idea of modeling relations as translations in embedding space. Given a triple $(h, r, t)$ stating that head entity $h$ is related to tail entity $t$ through relation $r$, TransE learns embeddings such that $h + r \approx t$. This simple geometric interpretation enables efficient training and inference while capturing meaningful relational structure.

RESCAL \cite{nickel2011} takes a tensor factorization approach, modeling each relation as a matrix that captures interactions between entity embeddings. The scoring function $f(h, r, t) = h^T M_r t$ computes how well a triple fits the learned model. This formulation directly connects to Tensor Logic's representation of relations as transformation matrices.

These methods demonstrated that relational knowledge could be captured through tensor operations, but did not explicitly connect this to logical inference. Tensor Logic provides the theoretical bridge by showing that logical rules and tensor operations are mathematically equivalent.

\subsection{The Tensor Logic Framework}

Domingos \cite{domingos2025} observes that a Datalog rule is an einsum over Boolean tensors with a step function applied elementwise to the result. Consider the ancestry rule:
\begin{equation}
\text{Ancestor}(x, z) \leftarrow \text{Ancestor}(x, y) \wedge \text{Parent}(y, z)
\end{equation}

Let $A$ and $P$ be Boolean matrices representing the Ancestor and Parent relations respectively, where $A_{xz} = 1$ if $x$ is an ancestor of $z$ and $P_{yz} = 1$ if $y$ is a parent of $z$. The rule can be written as:
\begin{equation}
A_{xz} = H\left(\sum_y A_{xy} \cdot P_{yz}\right) = H(A \times P)
\end{equation}
where $H$ is the Heaviside step function that converts positive values to 1 and non-positive values to 0. The step function is necessary because multiple derivation paths may contribute to the same conclusion, and we need Boolean output.

This equivalence extends to reasoning in embedding space. When entity embeddings are learned vectors rather than one-hot indicators, the same tensor operations perform analogical reasoning. Similar entities share similar embeddings, so inferences about one entity transfer to related entities with strength proportional to their similarity.

Relations can be represented as transformation matrices $M_r \in \mathbb{R}^{d \times d}$ that map subject embeddings to predicted object embeddings:
\begin{equation}
v_{\text{pred}} = v_{\text{subj}} \times M_r
\end{equation}

Truth is evaluated through the Gram matrix, comparing the predicted embedding to all entity embeddings via dot product:
\begin{equation}
\text{score}(o) = v_{\text{pred}} \cdot E[o]
\end{equation}

For compositional queries requiring multiple reasoning steps, relation matrices are chained:
\begin{equation}
v_{\text{pred}} = v_{\text{subj}} \times M_{r_1} \times M_{r_2} \times \cdots \times M_{r_n}
\end{equation}

This enables forward chaining in embedding space, where each relation application transforms the current state toward the final answer.

\section{Methods}

\subsection{Experiment 1: Symbolic Transitive Closure}

\subsubsection{Dataset}
We used the \texttt{BibleData-Person.csv} and \texttt{BibleData-PersonRelationship.csv} files from the Bible Data repository, which together contain 3,009 individuals and 5,450 relationship records covering family relationships, tribal affiliations, and other connections mentioned in the biblical text. This dataset is derived primarily from Protestant Christian canonical sources and traces lineages through to Jesus as recorded in the Gospels of Matthew and Luke. We note that Jewish genealogical traditions do not extend to Jesus, and Islamic traditions would include different figures (e.g., Muhammad). Our choice of dataset reflects data availability rather than theological preference.

We filtered the dataset to retain only parent-child relationships, specifically those labeled as father, mother, son, or daughter. We normalized edge direction so that all edges point from parent to child, enabling consistent matrix representation. After filtering, we obtained a directed graph with $N = 1{,}972$ nodes and $E = 1{,}727$ edges.

\subsubsection{Data Representation}

The Parent relation is represented as a sparse Boolean adjacency matrix $P \in \{0, 1\}^{N \times N}$:
\begin{equation}
P_{ij} = \begin{cases}
1 & \text{if person } i \text{ is a parent of person } j \\
0 & \text{otherwise}
\end{cases}
\end{equation}

This matrix representation directly corresponds to the extensional database in Datalog terminology. Each non-zero entry represents a ground fact in the knowledge base.

\subsubsection{Algorithm}

We implement forward chaining through iterative matrix multiplication. The Ancestor matrix $A$ is initialized to equal the Parent matrix $P$, establishing the base case that all parents are ancestors:
\begin{equation}
A^{(0)} = P
\end{equation}

Each iteration applies the recursive rule by computing new ancestor relationships and merging them with existing ones:
\begin{equation}
A^{(t+1)} = H\left(A^{(t)} + A^{(t)} \times P\right)
\end{equation}

The matrix product $A^{(t)} \times P$ computes one-step extensions of existing ancestor paths. Adding this to $A^{(t)}$ accumulates all discovered relationships. The Heaviside function $H$ converts the result back to Boolean values.

In NumPy, this is implemented as:
\begin{verbatim}
new = np.einsum('xy,yz->xz', Ancestor, Parent)
Ancestor = np.where((Ancestor + new) > 0, 1, 0)
\end{verbatim}

The algorithm terminates when an iteration discovers no new edges, indicating that the fixpoint has been reached. At this point, $A$ contains the complete transitive closure of the Parent relation.

\subsubsection{Verification}

We verify the correctness of the computed transitive closure through three formal checks:

\begin{enumerate}[leftmargin=*, itemsep=0.3em]
    \item \textbf{Containment:} $P \subseteq A$. All direct parent edges must appear in the transitive closure, since parents are ancestors by definition.
    
    \item \textbf{Closure:} $A \times P$ adds no new edges to $A$. If the transitive closure is complete, one additional step of chaining should not discover any new relationships.
    
    \item \textbf{Acyclicity:} $\text{diag}(A) = 0$. No person should be their own ancestor, which would indicate a cycle in the genealogy.
\end{enumerate}

We also perform interpretive validation by examining specific lineages and comparing them against known biblical genealogies. This provides qualitative confirmation that the tensor operations preserve the intended semantics.

\subsection{Experiment 2: Embedding Space Reasoning}

\subsubsection{Dataset}

We used the \texttt{countries.json} file from the mledoze/countries repository, which provides structured information about 250 countries including their capitals, regions, and various other attributes. From this data, we extracted two relations:

\begin{itemize}[leftmargin=*, itemsep=0.3em]
    \item \textbf{is\_capital\_of:} Pairs of the form (Capital, Country), e.g., (Tokyo, Japan)
    \item \textbf{is\_located\_in:} Pairs of the form (Country, Region), e.g., (Japan, Asia)
\end{itemize}

This yielded 489 unique entities comprising capitals, countries, and continental regions. The training set contains 490 facts consisting of 245 capital-country pairs and 245 country-region pairs.

Importantly, the training data contains no direct links between capitals and continents. Any query of the form ``What continent is [capital] in?'' requires compositional reasoning that chains the two learned relations.

\subsubsection{Model Architecture}

Following the Tensor Logic framework, we implement a model with learnable entity embeddings and relation transformation matrices.

\paragraph{Entity Embeddings.} Each entity is represented by a $d$-dimensional vector. The embedding matrix $E \in \mathbb{R}^{489 \times 64}$ is initialized using Xavier uniform initialization. Embeddings are normalized to unit length during training.

\paragraph{Relation Matrices.} Each relation is represented by a $d \times d$ transformation matrix. The relation tensor $M \in \mathbb{R}^{2 \times 64 \times 64}$ contains one matrix per relation, also initialized with Xavier uniform initialization.

\paragraph{Forward Pass.} Given a batch of subject indices $s$ and relation indices $r$, the model computes predicted object embeddings through tensor contraction:
\begin{equation}
v_{\text{pred}} = \text{einsum}(\texttt{'bi,bij->bj'}, E[s], M[r])
\end{equation}

\paragraph{Scoring.} Predicted embeddings are compared against all entity embeddings through dot product:
\begin{equation}
\text{scores} = v_{\text{pred}} \times E^T \in \mathbb{R}^{\text{batch} \times 489}
\end{equation}

\subsubsection{Training Configuration}

We train the model using the Adam optimizer with learning rate 0.005 for 500 epochs using full-batch training (490 facts) and embedding dimension $d = 64$. The loss function is cross-entropy with softmax normalization.

\subsubsection{Zero-Shot Compositional Inference}

At test time, we evaluate the model on queries that require chaining multiple relations. Given a capital city, we ask which continent it is located in. Since the training data contains no direct capital-continent links, answering these queries requires composing the learned relation matrices:
\begin{equation}
v_{\text{continent}} = v_{\text{city}} \times M_{\text{capital}} \times M_{\text{located}}
\end{equation}

\subsection{Experiment 3: Knowledge Graph Link Prediction and Compositional Reasoning}

\subsubsection{Dataset}
FB15k-237 \cite{toutanova2015} is a knowledge graph benchmark containing 14,541 entities and 237 relation types. We used the canonical \texttt{train.txt}, \texttt{valid.txt}, and \texttt{test.txt} splits, comprising 272,115 training triples, 17,535 validation triples, and 20,466 test triples. The dataset is derived from Freebase and represents realistic complexity found in large-scale knowledge graphs.

\subsubsection{Tensor Logic Superposition Construction}
Following Domingos's formulation, we construct each relation matrix as the superposition of tensor products of entity embeddings for all facts involving that relation:
\begin{equation}
R_r = E^\top A_r E
\end{equation}
where $E \in \mathbb{R}^{N \times d}$ is the learned entity embedding matrix ($N = 14{,}541$, $d = 256$) and $A_r \in \{0,1\}^{N \times N}$ is the sparse adjacency matrix for relation $r$, with $A_r[h,t] = 1$ if $(h, r, t)$ exists in the training set.

This construction is fully differentiable, and gradients flow through $E$ during training via:
\begin{equation}
R_r = E^\top A_r E = \sum_{(h,t) \in \text{facts}_r} e_h^\top e_t
\end{equation}
which matches the superposition-of-outer-products form in the Tensor Logic framework.

\subsubsection{Model Architecture}
Entity embeddings $E \in \mathbb{R}^{14541 \times 256}$ are initialized with Xavier uniform initialization and normalized to unit length during forward passes. Relation matrices $R_r$ are computed on-the-fly from the sparse adjacency matrices $A_r$ built from training facts.

For tail prediction (given head $h$ and relation $r$, predict tail $t$):
\begin{equation}
v_{\text{pred}} = \text{normalize}(e_h \cdot R_r), \quad \text{scores} = v_{\text{pred}} \cdot E^\top
\end{equation}

For head prediction (given tail $t$ and relation $r$, predict head $h$):
\begin{equation}
v_{\text{pred}} = \text{normalize}(e_t \cdot R_r^\top), \quad \text{scores} = v_{\text{pred}} \cdot E^\top
\end{equation}

\subsubsection{Training}
We train with bidirectional loss (both head and tail prediction) using AdamW optimizer (lr = 0.0005, weight decay = $10^{-5}$), batch size 1,024, temperature scaling $T = 0.1$, and gradient clipping at norm 1.0. Training runs for 50 epochs with validation every 10 epochs.

\subsubsection{Experiment 3A: Standard Link Prediction}
We evaluate on the canonical FB15k-237 test split using standard filtered ranking protocol. For each test triple $(h, r, t)$:
\begin{itemize}[leftmargin=*, itemsep=0.3em]
    \item Tail prediction: rank $t$ among all entities for query $(h, r, ?)$
    \item Head prediction: rank $h$ among all entities for query $(?, r, t)$
    \item Filtering: mask all other known true triples to avoid penalizing correct predictions
\end{itemize}
Model selection uses validation MRR; test set is evaluated once.

\subsubsection{Experiment 3B: Compositional Reasoning Benchmark}
To test whether matrix composition $R_{r_1} \cdot R_{r_2}$ captures multi-hop reasoning, we construct a train-only benchmark:
\begin{enumerate}[leftmargin=*, itemsep=0.3em]
    \item Extract 2-hop paths $(a \xrightarrow{r_1} b \xrightarrow{r_2} c)$ from TRAIN only
    \item Require a unique direct edge $(a, r_{\text{direct}}, c)$ also in TRAIN
    \item Remove these direct edges from training to prevent shortcut memorization
    \item At test time, predict $c$ via composition: $v_{\text{pred}} = e_a \cdot R_{r_1} \cdot R_{r_2}$
    \item Evaluate filtered ranking on query $(a, r_{\text{direct}}, ?)$
\end{enumerate}
This yields 1,000 validation paths and 1,000 test paths. The composition model trains on 270,115 triples (original 272,115 minus 2,000 removed shortcuts).

\subsubsection{Compositional Inference}
For each test path $(a, r_1, b, r_2, c, r_{\text{direct}})$:
\begin{equation}
v_{\text{pred}} = \text{normalize}(e_a \cdot R_{r_1} \cdot R_{r_2})
\end{equation}
Target $c$ is ranked among all 14,541 entities by dot-product score, with standard filtering on $(a, r_{\text{direct}})$.

\subsection{Computational Resources}
All experiments were conducted on a single NVIDIA H200 NVL GPU with 141GB HBM3e memory.

\textbf{Experiment 1 (Biblical Genealogy):} The iterative transitive closure computation converged in 74 iterations, completing in 163 seconds (2.7 minutes).

\textbf{Experiment 2 (Geographic Reasoning):} Training for 500 epochs completed in under 4 seconds.

\textbf{Experiment 3 (FB15k-237):} Standard link prediction training (50 epochs) required 66 minutes. Compositional reasoning training (50 epochs) required 64 minutes. Total GPU time for Experiment 3 was approximately 2.2 hours.
\section{Results}

\subsection{Experiment 1: Symbolic Transitive Closure}

\subsubsection{Convergence}

The iterative tensor approach successfully computed the complete transitive closure. Table \ref{tab:closure_stats} summarizes the key statistics.

\begin{table}[h]
\centering
\caption{Transitive Closure Statistics}
\label{tab:closure_stats}
\begin{tabular}{lr}
\toprule
\textbf{Metric} & \textbf{Value} \\
\midrule
Initial nodes & 1,972 \\
Initial edges (direct parent links) & 1,727 \\
Edges discovered in iteration 1 & 1,414 \\
Convergence iteration & 74 \\
Final edges (total ancestor links) & 33,945 \\
New edges discovered & 32,218 \\
\bottomrule
\end{tabular}
\end{table}

The algorithm required 74 iterations to reach the fixpoint. This convergence depth aligns with the expected generational span of biblical genealogies, which trace approximately 76 generations from Adam to Jesus according to the Gospel of Luke.

\subsubsection{Verification Results}

All three correctness checks passed:

\begin{enumerate}[leftmargin=*, itemsep=0.3em]
    \item \textbf{Containment ($P \subseteq A$):} All 1,727 direct parent edges appear in the final ancestor matrix.
    
    \item \textbf{Closure property:} Computing $A \times P$ and checking for entries not already in $A$ yields zero new edges.
    
    \item \textbf{Acyclicity:} The diagonal of $A$ contains only zeros. No individual is recorded as their own ancestor.
\end{enumerate}

\subsubsection{Lineage Analysis}

We examined specific lineages to validate that the computed relationships match known biblical genealogies.

\textbf{Adam} serves as a root node: 821 descendants, 0 ancestors. First-generation descendants (Abel, Cain, Seth) match Genesis.

\textbf{Abram} (Abraham): 698 descendants, 20 ancestors. The 10-generation ancestor chain (Terah $\rightarrow$ Nahor $\rightarrow$ Serug $\rightarrow$ Reu $\rightarrow$ Peleg $\rightarrow$ Eber $\rightarrow$ Shelah $\rightarrow$ Arpachshad $\rightarrow$ Shem $\rightarrow$ Noah) matches Genesis 11.

\subsection{Experiment 2: Embedding Space Reasoning}

\subsubsection{Training Convergence}

The model converged rapidly under cross-entropy loss. Initial loss was 6.19 (matching $\ln(489)$), decreasing to 0.003 by epoch 500, indicating near-perfect training accuracy.

\subsubsection{Zero-Shot Inference Results}

Table \ref{tab:zero_shot} shows results for compositional queries asking which continent each test city is located in. These queries were never seen during training.

\begin{table}[h]
\centering
\caption{Zero-Shot Compositional Inference Results}
\label{tab:zero_shot}
\begin{tabular}{llc}
\toprule
\textbf{City} & \textbf{Predicted Continent} & \textbf{Correct} \\
\midrule
Tokyo & Asia & \checkmark \\
Berlin & Europe & \checkmark \\
Cairo & Africa & \checkmark \\
Lima & Americas & \checkmark \\
Canberra & Oceania & \checkmark \\
New Delhi & Asia & \checkmark \\
King Edward Point & Antarctic & \checkmark \\
\bottomrule
\end{tabular}
\end{table}

All tested compositional queries returned correct answers, spanning all inhabited continents plus Antarctica. While this controlled experiment demonstrates that relation matrices compose as expected, Experiment 3 provides rigorous statistical validation at scale on FB15k-237.
\subsection{Experiment 3: Knowledge Graph Link Prediction and Compositional Reasoning}

\subsubsection{Experiment 3A: Standard Link Prediction}
Table \ref{tab:fb15k_lp} presents link prediction results on the canonical FB15k-237 test set.

\begin{table}[h]
\centering
\caption{FB15k-237 Standard Link Prediction (filtered, head + tail)}
\label{tab:fb15k_lp}
\begin{tabular}{lc}
\toprule
\textbf{Metric} & \textbf{Value} \\
\midrule
MRR     & 0.3068 \\
Hits@1  & 0.2215 \\
Hits@3  & 0.3368 \\
Hits@10 & 0.4766 \\
\bottomrule
\end{tabular}
\end{table}

The Tensor Logic superposition construction achieves competitive link prediction performance. For context, TransE achieves approximately 0.29 MRR on this benchmark, while more complex models like RotatE achieve 0.34. Recent state-of-the-art methods using graph attention mechanisms with reinforcement learning achieve MRR of 0.478 \cite{liu2026gac}. Our result of 0.3068 demonstrates that the $R_r = E^\top A_r E$ formulation effectively captures relational structure. We emphasize that our goal is to validate the mathematical correctness of Domingos's Tensor Logic construction, not to achieve state-of-the-art link prediction performance.

\subsubsection{Experiment 3B: Compositional Reasoning}
Table \ref{tab:fb15k_comp} presents compositional reasoning results on the held-out benchmark where direct edges were removed during training.

\begin{table}[h]
\centering
\caption{FB15k-237 Compositional Reasoning (1,000 test paths, direct edges removed)}
\label{tab:fb15k_comp}
\begin{tabular}{lc}
\toprule
\textbf{Metric} & \textbf{Value} \\
\midrule
MRR     & 0.3346 \\
Hits@1  & 0.2400 \\
Hits@3  & 0.3690 \\
Hits@10 & 0.5220 \\
\bottomrule
\end{tabular}
\end{table}

The composition benchmark achieves higher MRR (0.3346) than standard link prediction (0.3068). This is expected because:
\begin{itemize}[leftmargin=*, itemsep=0.3em]
    \item The benchmark is curated to include only paths where compositional structure exists
    \item Each test case has a guaranteed 2-hop path in the training graph
    \item Standard LP includes arbitrary queries, many without compositional shortcuts
\end{itemize}

Importantly, these compositional predictions succeed despite the model never observing the direct $(a, r_{\text{direct}}, c)$ edges during training. Inference succeeds purely through matrix composition $e_a \cdot R_{r_1} \cdot R_{r_2}$, validating Domingos's claim that ``reasoning in embedding space can be carried out by forward chaining over the embedded rules.''

\section{Discussion}

\subsection{Validation of Tensor Logic Claims}

Our experiments validate core claims from the Tensor Logic framework across three scales of complexity.

\paragraph{Claim 1: A Datalog rule is an einsum over Boolean tensors with a step function applied elementwise.}

Experiment 1 demonstrates this equivalence directly. The recursive Datalog rule for transitive closure is exactly implemented as \texttt{np.einsum('xy,yz->xz', A, P)} followed by the Heaviside step function. The algorithm converges to the correct fixpoint in 74 iterations, discovering all 33,945 ancestor relationships from 1,727 direct parent links.

\paragraph{Claim 2: Relation matrices compose for multi-hop reasoning.}
The geographic dataset (Experiment 2) provides a controlled setting demonstrating that learned matrices compose correctly for multi-hop inference. FB15k-237 (Experiment 3) shows that the superposition construction $R_r = E^\top A_r E$ scales to realistic knowledge graphs, achieving MRR 0.3068 on standard link prediction. More importantly, the compositional benchmark achieves MRR 0.3346 on queries where direct edges were removed, and inference succeeds purely through matrix composition $e_a \cdot R_{r_1} \cdot R_{r_2}$, validating forward chaining in embedding space.

\subsection{Experimental Progression}

The three experiments form a coherent validation strategy:

\begin{itemize}[leftmargin=*, itemsep=0.3em]
    \item \textbf{Symbolic (Exp 1):} Proves exact mathematical equivalence with Boolean tensors
    \item \textbf{Embedding space (Exp 2):} Demonstrates learned matrices compose perfectly in controlled settings
    \item \textbf{Knowledge graph (Exp 3):} Shows composition scales to realistic datasets with noise and statistical variance
\end{itemize}

Note that Experiments 2 and 3 test different instantiations of relation matrices within Tensor Logic: Experiment 2 uses directly learned transformation matrices $M_r$ as parameters, while Experiment 3 uses Domingos's superposition construction $R_r = E^\top A_r E$ derived from graph structure. Both approaches are consistent with the framework but validate complementary aspects, parameter learning versus structure-derived composition.

\subsection{Practical Implications}

\paragraph{Hardware Acceleration.} Tensor operations leverage optimized GPU kernels, enabling symbolic reasoning at neural network speeds.

\paragraph{Differentiability.} Matrix operations support end-to-end gradient-based learning. The Heaviside step function used in symbolic reasoning can be addressed through straight-through estimators when gradients are needed.

\paragraph{Compositional Generalization.} Learned matrices combine arbitrarily for zero-shot inference on novel queries without retraining. This capability emerged consistently across all random seeds in Experiment 3.

\paragraph{Transparency.} Explicit matrix representations provide more interpretability than black-box networks while maintaining learning capabilities.

\subsection{Limitations and Future Directions}

\paragraph{Scale.} While FB15k-237 demonstrates realistic scale (14K entities), web-scale graphs (millions of entities) require sparse representations and efficient approximation techniques.

\paragraph{Query Complexity.} Current experiments use two-hop chains. Queries with branching, negation, or aggregation would test deeper compositional capabilities.

\paragraph{Structure Learning.} Our experiments use fixed relation sets with learned parameters. Learning relation structure from data connects to inductive logic programming.

\paragraph{Extensions Beyond Datalog.} Our experiments validate Tensor Logic for Datalog-style reasoning, which represents a restricted fragment of symbolic AI. Broader symbolic reasoning includes first-order logic (with quantification), modal logics (possible worlds, belief states), probabilistic reasoning, and argumentation. Extending Tensor Logic to these richer formalisms remains an open challenge. Additionally, applying the framework to other neural architectures, such as CNNs for structured perception or attention mechanisms for sequence modeling could reveal whether the unification extends beyond knowledge graph reasoning.

\paragraph{State-of-the-Art Performance.} An open question is whether Tensor Logic can achieve state-of-the-art results on knowledge graph benchmarks through architectural extensions. The gap between our MRR (0.3068) and current best methods (0.478) likely reflects architectural differences; state-of-the-art systems use multi-layer attention, LSTM history encoding, and reinforcement learning for path search, rather than computational limitations. Whether such enhancements can be incorporated while preserving the logical interpretability of the Tensor Logic framework remains an important direction for future work.

\section{Conclusion}

We have empirically validated Tensor Logic across symbolic and neural reasoning tasks at three scales of complexity. Experiment 1 demonstrated exact equivalence between Datalog rules and tensor operations through transitive closure computation. Experiment 2 showed learned matrices compose correctly for zero-shot inference in a controlled setting. Experiment 3 scaled the Tensor Logic superposition construction to FB15k-237, achieving MRR 0.3068 on standard link prediction and MRR 0.3346 on compositional queries where direct edges were removed during training, demonstrating that matrix composition enables multi-hop inference. We note that our goal was to validate the Tensor Logic framework, not to achieve state-of-the-art performance; whether architectural extensions can close the gap with current best methods while preserving interpretability remains an open question.

These results support Domingos's thesis that Datalog rules and Einstein summation are fundamentally equivalent, differing only in atomic data types. While our validation is limited to Datalog-style reasoning rather than the full breadth of symbolic AI, it provides the first empirical evidence that Tensor Logic offers a principled foundation for unifying rule-based inference and neural learning within a single computational framework.

\section*{Acknowledgments}
We thank Pedro Domingos for proposing the Tensor Logic framework and for his encouragement and feedback during the development of this work. His vision of unifying symbolic and neural computation through tensor operations inspired these experiments. All errors and limitations in this paper are solely the responsibility of the authors.

\section*{Author Contributions}
Swapn Shah designed and implemented all experiments, conducted the empirical evaluations, and wrote the initial drafts of the manuscript. Wlodek Zadrozny provided research guidance, served as a sounding board for methodological decisions, and assisted with editing and refinement of the manuscript.

\section*{Code and Data Availability}
All implementation code and experimental data are available at:

\url{https://github.com/sshah100-clt/tensor_logic_implementation}

\noindent The FB15k-237 dataset can be obtained from:

\url{https://github.com/DeepGraphLearning/KnowledgeGraphEmbedding}

\noindent Biblical genealogy data from: \url{https://github.com/BradyStephenson/bible-data}

\noindent Geographic data from: \url{https://github.com/mledoze/countries}

\bibliographystyle{plain}

\end{document}